# Projected Subgradient Methods for Learning Sparse Gaussians

**John Duchi**
Computer Science Dept.
Stanford University
Stanford, CA 94305

**Stephen Gould**
Electrical Engineering Dept.
Stanford University
Stanford, CA 94305

**Daphne Koller**
Computer Science Dept.
Stanford University
Stanford, CA 94305

## Abstract

Gaussian Markov random fields (GMRFs) are useful in a broad range of applications. In this paper we tackle the problem of learning a sparse GMRF in a high-dimensional space. Our approach uses the $\ell_1$-norm as a regularization on the inverse covariance matrix. We utilize a novel projected gradient method, which is faster than previous methods in practice and equal to the best performing of these in asymptotic complexity. We also extend the $\ell_1$-regularized objective to the problem of sparsifying entire blocks within the inverse covariance matrix. Our methods generalize fairly easily to this case, while other methods do not. We demonstrate that our extensions give better generalization performance on two real domains—biological network analysis and a 2D-shape modeling image task.

## 1 Introduction

A key challenge in graphical models is learning model structure and variable independencies from data. We consider this problem in the context of Gaussian distributions defined over undirected graphs, or Gaussian Markov Random Fields (GMRFs). GMRFs are an important class of graphical models that have found applications in a wide range of areas including oceanography, image denoising, speech recognition, and FMRI (Willsky, 2002; Bilmes, 2000). In a GMRF, independencies—edges absent from the network—correspond to zero entries in the inverse covariance matrix $K$ (Lauritzen, 1996); that is, the GMRF has an edge between variables $i$ and $j$ if and only if $K_{ij} \neq 0$. Thus, learning a sparse GMRF corresponds precisely to learning an inverse covariance with many zero entries.

Sparsity in the inverse covariance matrix has a number of advantages. First, promoting sparsity when learning from limited data has been shown to produce robust models that generalize well to unseen data (Dempster, 1972). Second, the cost of both exact and approximate inference (e.g., belief propagation message passing (Malioutov et al., 2006)) depends strongly on the density of the GMRF structure. Finally, the number of parameters that need to be stored can be an important computational factor in some settings. These latter issues are particularly relevant in systems, such as speech recognition (Bilmes, 2000) or tracking (Bar-Shalom & Fortmann, 1988), that need to achieve real-time performance. Sparsity can also be beneficial in a knowledge discovery setting, where the structure of the learned model may provide insight into the relationships between the variables. For example, sparse Gaussian models have been successfully used to explore interactions between the genes in gene expression data (Dobra et al., 2004).

The idea of learning sparse GRMFs goes back to the work of Dempster (1972) where elements of the inverse covariance matrix are explicitly set to zero and the remaining parameters learned from data. Other early work used a greedy forward/backward search over edges in a Gaussian MRF that quickly became infeasible as $n$, the dimension of the problem, grew (Lauritzen, 1996). More recently, Banerjee et al. (2006) formulated the problem as one of optimizing a log-likelihood objective regularized with an $\ell_1$-norm penalty on the entries. This penalty is known (Tibshirani, 1996) to push parameters to zero, inducing sparsity. (See Section 2 for more details on this and other other approaches for learning sparse GMRFs.)

In this paper, we present a new algorithm for optimizing the $\ell_1$-penalized log-likelihood objective for Gaussian distributions. Our approach is based on projected gradient methods originally proposed by Levitin and Polyak (1966). Like other methods based on this objective, our method exploits the fact that the objective is tractable and convex, hence avoiding problems from greedy or heuristic searches. However, unlike the work of Banerjee et al. (2006), the complexity of our algorithm grows as $O(n^3)$ rather than $O(n^4)$ per iteration. For problems of high dimension, this order-of-magnitude reduction in complexity can be significant.

We also generalize the sparse inverse covariance problem to that of estimating an inverse covariance with block sparsity. This task corresponds to finding a Gaussian MRF in which certain groups of edges should all be penalized together. For example, in a GMRF over genes, we may want to jointly penalize any interaction between genes in two pathways; this penalty intuitively tries to reduce inter-pathway interactions, but once the two pathways are allowed to interact via one pair of genes, the penalty is removed for all other pairs. To our knowledge, this extension has not been considered elsewhere. Indeed, previous methods do not naturally handle this setting, whereas our method handles it easily, with no real increase in computational cost. We present results demonstrating the value of the block structure to two distinct applications: learning networks over genes from gene expression data, and learning models of the 2D shape of mammals.

## 2 Background and Related Work

As mentioned above, the literature on estimating covariance matrices from data has a long history. Our work starts from the formulation of Banerjee et al. (2006) which we review below. We then compare this approach to several other recent works.

### 2.1 $\ell_1$-Regularized Problem Formulation

We assume that we are given a dataset $\mathcal{D} = \{x[1], \ldots, x[m]\}$ in which the samples $x \in \mathcal{D}$ are drawn from some $n$-dimensional Gaussian distribution $\mathcal{N}(\mu, \Sigma)$. Given a Gaussian with mean $\mu$ and covariance $\Sigma$, the average log-likelihood (modulo unnecessary constants) for $\mathcal{D}$ with same mean can be written as

$$\log \det K - \text{tr}(\hat{\Sigma} K)$$

where $K = \Sigma^{-1}$ is the inverse covariance matrix for the model, and $\hat{\Sigma} = \frac{1}{m}\sum_{i=1}^{m}(x[i] - \mu)^T(x[i] - \mu)$ is the empirical covariance of the dataset $\mathcal{D}$. As we discussed, the sparsity structure in the matrix $K$ defines the structure of the GMRF corresponding to this Gaussian: a zero entry $K_{ij}$ corresponds precisely to the absence of an edge between the variables $i$ and $j$. For uniformity of exposition, we use the matrix terminology and notation in the remainder of our discussion.

Following the approach of Banerjee et al., we add sparsity-promoting $\ell_1$-penalty to the entries of the inverse covariance. We can now formulate the convex optimization problem for estimating the inverse covariance of our model:

$$\underset{K \succ 0}{\text{minimize}} \quad -\log \det(K) + \text{tr}(\hat{\Sigma} K) + \sum_{i,j} \lambda_{ij}|K_{ij}|, \quad (1)$$

where we use the notation $A \succ 0$ to indicate that the symmetric matrix $A$ is positive definite. Here $\lambda_{ij}$ controls the sparsity of the solution to (1). As we show in Lemma 1, so long as $\lambda_{ij} > 0$ for all $i \neq j$ (and $\lambda_{ii} \geq 0$) the solution to (1) is unique and positive definite. Furthermore, if $\lambda_{ij}$ is large enough, we will force all off-diagonal entries of $K$ to zero.

It will be instructive now and for later development in the paper to take the dual of (1). We introduce an auxiliary variable $Z = K$ and its associated dual variable $W \in \mathbb{R}^{n \times n}$, leading to the Lagrangian

$$\begin{aligned}\mathcal{L}(K, Z, W) &= -\log\det(K) + \text{tr}(\hat{\Sigma} K) \\ &\quad + \sum_{i,j} \lambda_{ij}|Z_{ij}| + \text{tr}(W(K - Z)).\end{aligned}$$

The above Lagrangian is separable into terms involving $K$ and terms involving $Z$, which enables us to find the dual. For the terms involving $Z$ we have that

$$\inf_Z \sum_{ij} \lambda_{ij}|Z_{ij}| - \text{tr}(WZ) = \begin{cases} 0 & \text{if } |W_{ij}| \leq \lambda_{ij} \\ -\infty & \text{otherwise.} \end{cases}$$

The infimum over $K$ for the remaining terms can be found by using $\nabla_K \log\det(K) = K^{-1}$ (and assuming that $\hat{\Sigma} + W \succ 0$) to give

$$\inf_K [-\log\det(K) + \text{tr}((\hat{\Sigma}+W)K)] = \log\det(\hat{\Sigma}+W) + n$$

Combining, we get the following dual problem for (1):

$$\begin{aligned}\text{maximize} \quad & \log\det(\hat{\Sigma} + W) \\ \text{subject to} \quad & |W_{ij}| \leq \lambda_{ij} \;\; \forall i, j.\end{aligned} \quad (2)$$

The constraint that $\hat{\Sigma} + W \succ 0$ is implicit in the objective because of the convention that $\log\det(X) = -\infty$ when $X \not\succ 0$. We denote the set $\{W : |W_{ij}| \leq \lambda_{ij} \;\forall i,j\}$ by $B_\lambda$, which is a box constraint indexed by the tuple $\lambda = (\lambda_{ij})$. The duality gap given a dual feasible point $W \in B_\lambda$ (and hence the primal point $K = (\hat{\Sigma} + W)^{-1}$) is

$$\eta = \text{tr}(\hat{\Sigma} K) + \sum_{ij} \lambda_{ij}|K_{ij}| - n.$$

Based on the dual problem, we prove the following lemma, which guarantees a solution to problem (1):

**Lemma 1.** *With probability 1, so long as $\lambda_{ij} > 0$ for all $i \neq j$ and $\lambda_{ii} \geq 0$, the minimization problem (1) is bounded below and has a unique optimal point $K^\star$.*

**Proof** We consider the case when $\lambda_{ii} = 0$ as Banerjee et al. (2006) already consider the case when $\lambda_{ii} > 0$. First, with probability 1, $\hat{\Sigma}$ has non-zero diagonals because it was generated from a set of data samples. If there is a dual feasible point for (2), i.e., some $W$ such that $\text{diag}(W) = 0$ and $|W_{ij}| \leq \lambda_{ij}$, the dual problem has a non-infinite value, and so the primal is bounded below. Further, if we can find a dual-feasible $W$ such that $|W_{ij}| < \lambda_{ij}$ (i.e. $W$ is in the relative interior of the domain), then Slater's constraint qualification (Boyd & Vandenberghe, 2004) guarantees a primal-dual optimal pair $K^\star = (\hat{\Sigma} + W^\star)^{-1}$ with zero duality gap. So we simply show that a $W$ in the relative interior of the domain exists.

Consider the matrix $D = \text{diag}(\hat{\Sigma}) \succ 0$. For $\alpha \in [0, 1)$ we have $\alpha \hat{\Sigma} + (1-\alpha)D \succ 0$. But, for any $\alpha$, we have that $\text{diag}(\alpha\hat{\Sigma} + (1-\alpha)D) = \text{diag}(\hat{\Sigma})$. Thus, we choose

$$W = \alpha\hat{\Sigma} + (1-\alpha)\text{diag}(\hat{\Sigma}) - \hat{\Sigma}, \quad (3)$$

which has zeros on the diagonal, and for $\alpha$ close enough to 1 will have $|W_{ij}| < \lambda_{ij}$. Further,

$$\hat{\Sigma} + W = \alpha\hat{\Sigma} + (1-\alpha)\operatorname{diag}(\hat{\Sigma}) \succ 0.$$

Choosing $W$ as in (3) gives a dual feasible point in the relative interior of the domain. $W^\star$ is unique by the strong convexity of $\log\det$ over the positive definite cone. □

If we can find the maximizing $W^\star$ for (2), we can also easily calculate the sparsity for $K^\star = (\hat{\Sigma} + W^\star)^{-1}$ by the complementarity conditions on $W_{ij}$ and $K_{ij}$ (Boyd & Vandenberghe, 2004). Specifically, if $|W^\star_{ij}| < \lambda_{ij}$, then $K^\star_{ij} = 0$. Lastly, as shown in the analysis of the so-called maximum determinant completion problem (see Boyd and Vandenberghe (2004)), if we constrain only the diagonal elements of $W$ and do not constrain its off-diagonals, then maximizing $\log\det(\hat{\Sigma}+W)$ will drive the off-diagonal entries of $K$ to 0; as such, choosing $\lambda_{ij} = |\hat{\Sigma}_{ij}|$ forces complete sparsity in $K$ and we find an MRF with no edges, i.e., all the variables are independent. Thus, when solving (1), we have a natural upper bound on $\lambda$ beyond which no additional sparsity can be gained.[1]

## 2.2 Prior Work

One recent approach for learning sparse GMRFs is to focus on learning the set of neighbors for a particular variable by regressing that variable against the remaining variables in the network with an $\ell_1$-penalty to promote sparsity (Meinshausen & Bühlmann, 2006). An alternative approach, first presented by Banerjee et al. (2006), is to solve the optimization problem (2). It is convex and can be solved using interior point methods in $O(n^6 \log(1/\varepsilon))$ time to a desired accuracy $\varepsilon$, however, this becomes infeasible for even moderate $n$. Banerjee et al.'s approach was to solve problem (2) iteratively, column by column, as a sequence of QPs. That is, they define $U = \hat{\Sigma} + W$ and partition $U$ and $\hat{\Sigma}$ as

$$U = \begin{bmatrix} U_{11} & u_{12} \\ u_{12}^T & u_{22} \end{bmatrix} \quad \hat{\Sigma} = \begin{bmatrix} S_{11} & s_{12} \\ s_{12}^T & s_{22} \end{bmatrix}.$$

They then solve a sequence of box-constrained QPs (swapping rows of $U$ to get all the rows), each iteration setting $\hat{u}_{12} = \operatorname{argmin}_y \{y^T U_{11}^{-1} y \mid \|y - s_{12}\| \le \lambda\}$, updating $U$ and $W$ appropriately. This method takes $O(Tn^4)$ time, where $T$ is the number of passes through the columns of the matrix $U$.

An alternative to the box-constrained QP above (its dual) is to minimize a sequence of re-weighted LASSO problems. In work developed in parallel and independently of our paper, Friedman et al. (2007) use this to solve problem (2). Their method enjoys better performance (around $O(Tn^3)$, though no analysis is given) than Banerjee et al.'s, which is similar to the methods we develop in our paper. They, however, explicitly rely on $\lambda_{ii}$ being positive to get a feasible starting point when the number of samples is less than the dimension, which we do not.

We have already mentioned Meinshausen and Bühlmann's work (2006) regressing single variables against one another. The advantage of their approach is in its efficiency, as solving $n$ regressions each with $n$ variables can be done very quickly, i.e., in the time it takes to do one pass over all the columns with LASSO. However, they do not directly obtain a maximum likelihood estimate from their method, only the structure of the graph.

The SPICE algorithm of Rothman et al. (2007) solves (1) when $\lambda_{ii} = 0$ (i.e., applying an $\ell_1$-penalty only to off diagonal elements of the inverse covariance). The advantages of this penalty are many: Rothman et al. prove that such a setting of the regularization terms is consistent so that the solution to (1) approaches the true inverse covariance as the number of data samples increases, and they give experiments showing that not penalizing the diagonal consistently learns model structure more accurately than does regularizing each entry of $K$. Further, Meinshausen (2005) gives simple conditions under which penalizing the on-diagonal entries of $K$ in (1) gives the *wrong* inverse covariance, even in the limit of infinite data. These results underscore the importance of Lemma 1, which gives a dual-feasible starting point even when $\lambda_{ii} = 0$. To our knowledge, this has not been noted before this work.

The disadvantage of the SPICE algorithm is that, like Banerjee et al.'s algorithm, it performs coordinate-wise updates (though it uses the columns of the Cholesky decomposition of $K$). It repeatedly iterates through all the columns, solving $n$ $O(n^3)$ regressions at every step, giving it a time complexity again of $O(Tn^4)$. Further, Rothman et al.'s algorithm has no explicit way to check convergence, as it does not generate dual-feasible points (which our algorithms do), and they rely on truncation of near-zero values rather than explicit conditions for sparsity in the inverse covariance (such as complementarity of primal-dual variables). The algorithm they propose is also somewhat complicated.

## 3 Projected Gradient Method

We propose a projected gradient method for solving the dual problem (2). Projected gradient algorithms minimize an objective $f(x)$ subject to the constraint that $x \in S$ for some convex set $S$.[2] They do this by iteratively updating

$$x := \Pi_S(x + t\nabla f(x))$$

where $t$ is a step size and $\Pi_S(z) = \operatorname{argmin}_y\{\|z-y\|_2 \mid y \in S\}$ is the Euclidean projection onto set $S$ (Bertsekas, 1976). First order projected gradient algorithms are effective when second order methods are infeasible because of the dimension of the problem. As the dimensions of our problems

---

[1] Presumably one could regularize the diagonals further, though this seems of little practical use.

[2] In our case $S$ is the set of $n \times n$ matrices in the box $B_\lambda$.

**Algorithm 1** Maximize $\log\det(\hat{\Sigma} + W)$ subject to $W \in B_\lambda = \{W \mid |W_{ij}| \leq \lambda_{ij}\}$. Given empirical covariance $\hat{\Sigma} \succeq 0$, $\lambda_{ij}$, duality gap stopping criterion $\epsilon$, and $W$ such that $W \in B_\lambda$, $\hat{\Sigma} + W \succ 0$, and $W_{ii} = \lambda_{ii}$.

1: **repeat**
2:     *Compute unconstrained gradient*
3:         $G := (\hat{\Sigma} + W)^{-1}$
4:     *Zero components of gradient which would result in constraint violation*
5:         $G_{ii} := 0$
6:         $G_{ij} := 0$ for all $W_{ij} = \lambda_{ij}$ and $G_{ij} > 0$
7:         $G_{ij} := 0$ for all $W_{ij} = -\lambda_{ij}$ and $G_{ij} < 0$
8:     *Perform line search*
9:         $t :\approx \mathrm{argmax}_t \log\det(\hat{\Sigma} + \Pi_{B_\lambda}(W + tG))$
10:    *Update and project*
11:        $W := \Pi_{B_\lambda}(W + tG)$
12:        $K := (\hat{\Sigma} + W)^{-1}$
13:    *Compute duality gap*
14:        $\eta = \mathrm{tr}(\hat{\Sigma}K) + \sum_{ij} \lambda_{ij}|K_{ij}| - n$
15: **until** $(\eta < \epsilon)$ or maximum iterations exceeded
16: **return** $K$

---

**Algorithm 2** Line search to find feasible $t$ for $\log\det(\hat{\Sigma} + W + tG)$ given $f_0 = \log\det(\hat{\Sigma} + W)$.

1:  $t := \frac{\mathrm{tr}((\hat{\Sigma}+W)^{-1}G)}{\mathrm{tr}((\hat{\Sigma}+W)^{-1}G(\hat{\Sigma}+W)^{-1}G)}$
2: **while** $\log\det\left(\hat{\Sigma} + \Pi_{B_\lambda}(W + tG)\right) \leq f_0$ **do**
3:    $t := t/2$
4: **end while**

---

often exceed $n = 1000$, giving more than 500,000 different parameters, this makes projected gradient methods a reasonable choice.

The projected gradient method for our problem is shown in Algorithm 1. The unconstrained gradient of the dual objective function (2) is $G = (\hat{\Sigma} + W)^{-1}$. We perform a line search to find the step size $t$ that approximately gives the greatest increase to the objective. This search needs to guarantee that the estimated covariance $\hat{\Sigma} + W$ is positive definite, which it does because we assume that $\log\det X = -\infty$ for $X \not\succ 0$. We can guarantee that the initial $\hat{\Sigma} + W \succ 0$, because $\hat{\Sigma} \succeq 0$, and we can simply initialize $W_{ii} = \lambda_{ii}$ and the rest of $W$ via Lemma 1 (this initialization is optimal for the diagonal elements, so we do not modify them through the course of the algorithm). The projection of the gradient appears in two places. First, we immediately zero out some entries of the gradient when $W_{ij}$ is at the boundary and $G_{ij}$ would push $W_{ij}$ outside $B_\lambda$. Second, during the line search, in each step we project the gradient onto the box-constraint $|W_{ij}| \leq \lambda_{ij}$ via the operation $\Pi_{B_\lambda}(W + tG)$, which simply sets any entry $> \lambda_{ij}$ to $\lambda_{ij}$ (and likewise for $-\lambda_{ij}$).

For the simple box-constrained projections, because we can immediately zero out many entries of the gradient $G$ and still have a descent direction (see lines 5 through 7 of Algorithm 1, which are not strictly necessary but improve the performance of the line search), a simple heuristic line search based on the second-order approximation to $\log\det$ performs very well. The second order expansion of the log-determinant function around a point $X$ (Boyd & Vandenberghe, 2004) is given by $\log\det(X + \Delta X) \approx \log\det(X) + \mathrm{tr}(X^{-1}\Delta X) - \frac{1}{2}\mathrm{tr}(X^{-1}\Delta X X^{-1}\Delta X)$. Thus, given the descent direction $G$, we approximate $\log\det(\hat{\Sigma} + W + tG)$ by

$$\log\det(\hat{\Sigma} + W) + t\,\mathrm{tr}((\hat{\Sigma}+W)^{-1}G)$$
$$- \tfrac{1}{2}t^2\,\mathrm{tr}((\hat{\Sigma}+W)^{-1}G(\hat{\Sigma}+W)^{-1}G)$$

and perform the line search of Algorithm 2. Convergence is guaranteed because the iterates for $W$ form a sequence in a compact space $B_\lambda$ and $\log\det(\hat{\Sigma}+W)$ is always increasing. If the line search cannot return a satisfying $t$, then no improvement can be made in the projected descent direction, so standard arguments by KKT conditions for a differentiable convex function guarantee that we have reached the optimum.

## 4 Structure Extensions

We can extend the basic problem of (1) to cases in which we are interested in sparsity not just between single variables but between entire blocks of variables. In many problem domains, variables can be naturally grouped into blocks. For example, we might try to model a 2D shape made up of articulated objects (such as the outline of an animal) in which we want to regularize interactions between object parts (such as legs, body, head, and tail). Landmarks along the contour of an animal's head can naturally be grouped together, as these landmarks move collectively as the animal moves through different articulated forms. In modeling gene networks, we may want to encode the intuition that interactions happen at the level of pathways, i.e., either two pathways interact, in which case multiple genes can be involved, or they do not interact at all. Block regularization might also arise in the context of learning multi-resolution models constrained so that only variables within specific resolution levels interact (Willsky, 2002).

In the block regularization case, we let the entries in our inverse covariance matrix be divided into $p < n^2$ disjoint subsets $S_1, \ldots, S_p$ (the disjointness assumption is essential to our method, as we see below). We can find the dependencies between the subsets by solving the following block $\ell_1$-regularized log likelihood problem:

$$\begin{aligned}\text{minimize} \quad &-\log\det(K) + \mathrm{tr}(\hat{\Sigma}K) \\ &+ \sum_k \lambda_k \max\{|K_{ij}| : (i,j) \in S_k\}.\end{aligned} \quad (4)$$

The subset $S_k$ encodes a set of interactions that, as soon as one of the interactions in the set exists, there is no reason to penalize any of the other interactions. For example, in the 2D shape model, the subsets $S_k$ may be constructed as $S_{qr} = \mathcal{B}_q \times \mathcal{B}_r$ where the $\mathcal{B}_q$ and $\mathcal{B}_r$ represent the set of variables belonging to the $q^{th}$ and $r^{th}$ articulated body part, respectively.

We can perform a similar derivation to the one in Section 2.1 to find the dual problem for (4):

$$\text{maximize} \quad \log \det(\hat{\Sigma} + W) \quad (5)$$
$$\text{subject to} \quad \sum_{(i,j) \in S_k} |W_{ij}| \leq \lambda_k, \quad \text{for } k = 1, \ldots, p.$$

To apply a projected gradient method to (5), we need to project to the constraints $\sum_{(i,j) \in S_k} |W_{ij}| \leq \lambda_k$. Recent work by Duchi et al. (2008) provides such a method, giving a randomized linear time algorithm to project a vector $w$ to the constraint $\|x\|_1 \leq C$. The sets $S_k$ must be disjoint so that we can project to each constraint in (5) independently (and hence efficiently). We refer the reader to their paper for a description of the actual projection algorithm.

Given the linear time algorithm to project to an $\ell_1$ constraint, we can develop an $O(n^2)$ expected time (the same efficiency as the projection to $B_\lambda$ of earlier methods) method to project to the feasible set $S = \{W \mid \sum_{(i,j) \in S_k} |W_{ij}| \leq \lambda_k, k = 1, \ldots, p\}$. The method simply iterates through each of the blocks $S_k$, projecting $W_{ij} : (i,j) \in S_k$ to the constraint that $\sum_{(i,j) \in S_k} |W_{ij}| \leq \lambda_k$ to get $W \in S$. With this as a building block, Algorithm 3 maximizes the dual (5). Because the constraint set is more complicated, making it difficult to make the second order expansion of $\log \det$ accurate for the projected direction, the algorithm uses Armijo-like line searches as described by Bertsekas (1976) rather than the simpler line search of Algorithm 1. If we let $g = \nabla f(x)$, the Armijo line search for a projected gradient maximization method with function $f$ returns the first $t$ such that

$$f(\Pi_S(x + tg)) \geq f(x) + \alpha \nabla f(x)^T (\Pi_S(x + tg) - x)$$

where $t$ is initialized to 1 and decreased by a multiplier $\beta < 1$ every time the above condition is not satisfied. In our algorithm $t$ is adaptively chosen to make the line search converge more quickly (see line 22 of Alg. 3), and setting $0 < \alpha < 1$ guarantees convergence of the method.

Previous methods developed for solving the penalized maximum-likelihood problem for Gaussians are unable to handle the block $\ell_1$-penalties of (4). Effectively, batch steps are needed to handle the more long range block constraints; the coordinate based methods of Banerjee et al. (2006), Friedman et al. (2007), and Rothman et al. (2007) cannot account for more global constraints that tie parameters in arbitrary columns and rows. Our projected gradient methods, however, extend to the case of block penalties, with ease. Further, the added complexity is negligible, as the entire projection step is still $O(n^2)$, while the expensive $O(n^3)$ step is computing the gradient.

**Algorithm 3** Solves (5) given an initial $W$ such that $\hat{\Sigma} + W \succ 0$ and constants $\beta$ and $\alpha$, $0 < \beta < 1$ and $0 < \alpha < 1$.
1: $t = 1$
2: **repeat**
3:    *Compute unconstrained gradient*
4:      $G = (\hat{\Sigma} + W)^{-1}$
5:    *Compute direction of step*
6:      $D = \Pi_S(W + tG) - W$
7:    *Compute initial and next objective values*
8:      $f_0 = \log \det(\hat{\Sigma} + W)$
9:      $f_t = \log \det(\hat{\Sigma} + \Pi_S(W + tG))$
10:   *Perform backtracking line search*
11:   **while** $f_t < f_0 + \alpha \operatorname{tr}(DG)$ **do**
12:     *Decrease $t$, recalculate direction and objective*
13:      $t = \beta \cdot t$
14:      $D = \Pi_S(W + tG) - W$
15:      $f_t = \log \det(\hat{\Sigma} + \Pi_S(W + tG))$
16:   **end while**
17:   *Compute next points and duality gap*
18:     $W = \Pi_S(W + tG)$
19:     $K = (\hat{\Sigma} + W)^{-1}$
20:     $\eta = \operatorname{tr}(\hat{\Sigma}K) + \sum_k \lambda_k \|K_{ij} : (i,j) \in S_k\|_\infty - n$
21:   *Increase $t$ slightly*
22:     $t = t/\beta$
23: **until** $\eta < \epsilon$

## 5 Experimental Results

In this section, we describe our experimental results. We performed experiments both on synthetic and real data, gathering timing information as well as calculating log-likelihood on test data, validating the usefulness of sparse estimators. This validation seems to have been notably absent from much of the literature on sparse inverse covariance selection.

### 5.1 Timing Results

We ran a series of timing experiments for the original problem from equations (1) and (2) comparing our approach to that of Banerjee et al. (2006) and Friedman et al. (2007). To generate data for our timing experiments, we constructed random $n \times n$ sparse inverse covariance matrices with roughly 20 edges per node. For each such matrix, we generated $n/3$ samples and used them to construct the empirical covariance $\hat{\Sigma}$. We selected $\lambda$ for the penalty in (1) so that at solution, the inverse covariance $K$ had (approximately) the correct number of edges. Our timing experiments were run on a computer with a 1.7Ghz Intel Xeon 32-bit processor and 1.96GB of RAM. The run times to achieve a duality gap of $\epsilon = 0.1$ are plotted in Fig. 1, which shows CPU time versus problem size compared to Banerjee et al.'s and Friedman et al.'s column-wise coordinate ascent algorithms. The results show that our projected subgradient method outperforms Banerjee et al.'s column-wise ascent by one to two orders of magnitude on these tests. Our method seems to find a solution in roughly half to two-thirds the time that Friedman et al.'s does, though their method seems to more easily be optimized when the primal matrix $K$ becomes sparse. Further, in our experiments when $\hat{\Sigma}$ was not full rank, Friedman et al.'s method seemed to have more difficulty maintaining a positive definite solution than did ours, which guarantees positive definiteness of $\hat{\Sigma} + W$ throughout. We note that due to the slightly more complicated projection, the run times of Alg. 3 are roughly twice those of Alg. 1.

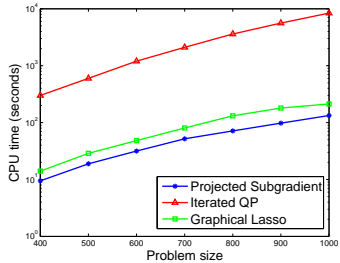

**Figure 1:** Run times in log-scale of the method presented in this paper versus Banerjee et al.'s (Iterated QP) and Friedman et al.'s (Graphical Lasso).

### 5.2 Synthetic Log Likelihoods

To test the impact of learning sparse structures, we compared our $\ell_1$-penalized inverse covariance estimation from (1) to Tikhonov regularized covariance matrices, i.e., selecting the covariance matrix to be $\Sigma = \hat{\Sigma} + \nu I$ for some $\nu > 0$, which guarantees the positive-definiteness of $\Sigma$. To do this, we randomly generated 20 inverse covariance matrices with 50% sparsity and dimension $n = 60$. From each of these, we generated 30 samples for a training set (defining a covariance $\hat{\Sigma}$) and 30 samples for a test set. We then varied $\nu$ to compute the best *test set* log-likelihood achieved by Tikhonov regularized covariance across all $\nu$'s. Fig. 2(a) shows this best-case test log-likelihood for the Tikhonov regularized covariance versus the log-likelihood for sparse covariances matrices output by Algorithm 1 as we sweep the penalty parameter $\lambda$ from $\max(|\hat{\Sigma}_{ij}|)$ to $10^{-3}$. The results show that, for appropriate levels of sparsity, $\ell_1$-regularized covariance estimation outperforms simple regularized estimates of the full covariance.

### 5.3 Mammal 2D Shape Models

Moving on to real data, we aimed to study whether our block $\ell_1$ approach achieves better generalization than other approaches for learning the model. In our first application, we consider a two-dimensional shape classification task. Here, we have a series of 60 landmarks, each an $(x, y)$ point in 2D, defining the outline of a mammal as a vector in $\mathbb{R}^{120}$. The mammals are from six classes: bison, deer, elephants, giraffes, llamas, and rhinos. There are on average 42 examples from each class, each a hand-labeled outline from a real image. Our task is to model each animal's outline as a Gaussian distribution, learning a mean $\mu^{(c)}$ and covariance $\Sigma^{(c)}$ for each of the six animal classes $c$.

We used and compared the standard sparsifying objective

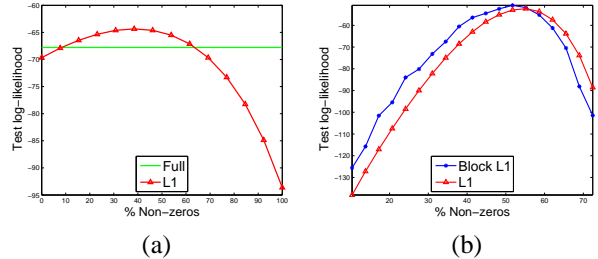

(a)      (b)

**Figure 2:** Log-likelihood results: (a) for synthetic data, comparing $\ell_1$-penalized inverse covariance to (best-case) full covariance; (b) for 2D shape-recognition test data, comparing the block versus the non-block method.

(1) and the block-penalized objective (4) to learn sparse inverse covariances $K^{(c)}$. For the block penalties, we manually chose the articulated body parts $\mathcal{B}_q$ as the head, neck, stomach, each leg, back, and tail. The blocks $S_k$ were then $S_k = \mathcal{B}_q \times \mathcal{B}_r$ for all pairs $q$ and $r$ (including $q = r$), and the penalty $\lambda_k$ for $S_k$ was set so that $\lambda_k \propto |S_k|$.[3]

We measure the performance of the competing methods in two ways: by test set log-likelihood and by performance on a classification task. In the log-likelihood case, we performed five-fold cross validation of the training data and swept the $\lambda$ penalties for both (1) and (4), ranging from full-sparsity to no-sparsity solutions for the inverse covariance $K$. We set the mean $\mu^{(c)}$ for each class simply as the training set mean. As a baseline, we also chose a "full" covariance matrix $\hat{\Sigma} + \nu I$ for each class using the Tikhonov regularization technique described above. As above, we selected $\nu$ to give the best test log-likelihood for each class in our cross-validation procedure. The block sparsity led to moderate improvements in the test likelihood, which can be seen in Fig. 2(b); we note that we do not plot the full covariance result, as its best average log-likelihood on test data is -173.22, which is off the bottom of the plot. We see that $\ell_1$-regularization significantly improves generalization, and that our block approach provides yet an additional improvement.

Other benefits of the block structured regularization can be seen in a classification task. In this case, for each of the three methods (sparse, block sparse, and Tikhonov regularized) and for each class we chose the inverse covariance $K^{(c)}$ that maximized the Gaussian log-likelihood on a held-out validation set. The task is to classify examples from a test set, that is, to assign a label $\hat{c}$ to a given vector $v \in \mathbb{R}^{120}$ from the test set. We assign the label for $v$ simply as the class that has the maximum likelihood for $v$:

$$\hat{c} = \underset{c}{\operatorname{argmax}} \left\{ \log \det K^{(c)} - (v - \mu^{(c)})^T K^{(c)} (v - \mu^{(c)}) \right\}.$$

The false positive and false negative error rates over ten different testing runs for each class are in Table 1. We see that the standard $\ell_1$-regularized objective almost always outperforms the full inverse covariance, and that the block $\ell_1$-regularized likelihood consistently and significantly outperforms both other methods.

---

[3] Having the penalties on the blocks correspond to their sizes overcomes the tendency for larger blocks to have edges since they can more easily affect the log-likelihood.

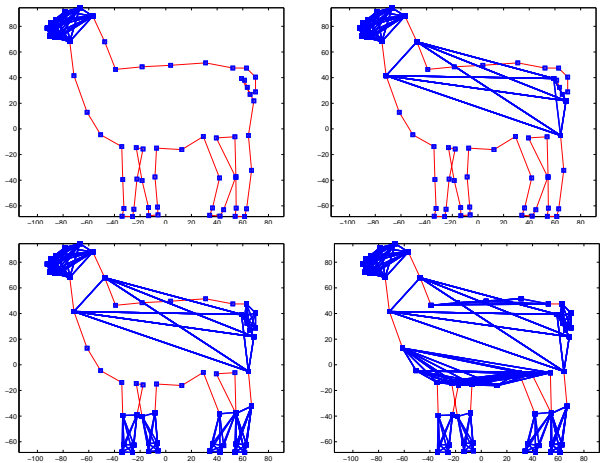

**Figure 3:** Different sparsity structures learned on blocked llama shape data.

**Table 1:** Classification Error Rates on Animals

| NEG | Bison | Deer | Elephant | Giraffe | Llama | Rhino |
|---|---|---|---|---|---|---|
| $\ell_1$ | 5.6% | 2.2% | 1.3% | 5.0% | 40.0% | **3.3%** |
| Block | **1.1%** | **0%** | **0%** | **3.0%** | **35.5%** | **3.3 %** |
| Full | 10.0% | 1.1% | 2.5% | 6.0% | 41.1% | 10.0% |
| POS | Bison | Deer | Elephant | Giraffe | Llama | Rhino |
| $\ell_1$ | 1.2% | 35.3% | **1.2%** | **0%** | 3.6% | **0%** |
| Block | **1.1%** | **29.1%** | **1.2%** | **0%** | **0%** | **0%** |
| Full | 2.4% | 40.3% | 1.3% | **0%** | 1.9% | **0%** |

The block regularized covariance selection can also be visualized as in Fig. 3. In the figure, we display the sparsity (i.e. the edges in the associated MRF) of the block inverse covariance learned for a llama outline as we vary the penalty term. We can see that generally, the articulated parts (legs, head, tail, body) have edges between themselves before edges between parts; intuitively, the relationships within articulated parts, such as the head or the legs, capture the shape of the part and are likely to be more informative for density estimation of the shape distribution than long range interactions such as the head's position in relation to the back right leg.

### 5.4 Gene Expression Data

We considered a data set that measures the mRNA expression levels of the 6152 genes in *S. cerevisiae* (baker's yeast), measured under various environmental stress conditions (Gasch et al., 2000). The expression level of the genes can be modeled as random variables and each experiment as a data sample. Some of the expression data was missing and we used a standard nearest-neighbor method to impute the missing values (Troyanskaya et al., 2001). We restricted the dataset to the 667 genes involved in known metabolic pathways (Förster et al., 2003). We preprocess the data so that each variable is zero mean and unit variance across the dataset.

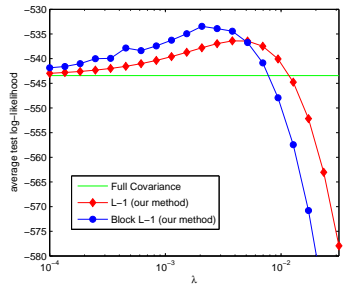

**Figure 4:** Average test log-likelihood on Gasch dataset for $\ell_1$ (red diamonds) and block $\ell_1$ (blue circles) regularization of inverse covariance matrix, compared to the full-covariance baseline (horizontal green line).

For the block experiments we group genes into 86 disjoint sets $\mathcal{B}_q$ that correspond to known metabolic pathways (Förster et al., 2003).[4] We then construct the edge subsets as follows. We have one block $S_{qr}$ for each pair of pathways $q \neq r$, which contains all the covariance entries for $\mathcal{B}_q \times \mathcal{B}_r$. We then have a separate block $S_{ij}$ for each pair of genes $i, j$ in the same pathway. That is, we apply block $\ell_1$ regularization between groups of genes in different blocks and standard $\ell_1$ regularization between individual genes within the same block. For each block $S_k$ we again set $\lambda_k = \lambda \cdot |S_k|$ where $\lambda$ is a shared regularization parameter.

We conducted 5-fold cross-validation on the dataset. On each fold we use the training set to estimate the covariance matrix and compute the log-likelihood for each instance in the test set. We report the average log-likelihood over the 5 folds. We baseline our method by first estimating the full covariance matrix. Since the number of data samples (174) is significantly smaller than the number of variables (667), we again use cross-validated Tikhonov regularization as previously described to give an estimate for $\Sigma$. We compare our results against the best (highest log-likelihood on test) Tikhonov regularized covariance.

Fig. 4 shows the performance of our methods compared to the baseline. The results clearly illustrate that the sparsity inducing $\ell_1$-regularized objective outperforms the baseline, while the block $\ell_1$-regularized penalties give further benefits in terms of log-likelihood.

## 6 Conclusion

In this work, we have presented new methods for finding sparse inverse covariance matrices and thereby for selecting edge structures for Gaussian MRFs. The methods we present are significantly faster than prior work, and they generalize straightforwardly to learning more complicated structures than has been previously possible. We also provide compelling experimental results on two real-world data sets demonstrating the benefits of this approach for both density estimation and classification.

---
[4]A small number of genes that participate in more than one pathway were placed into individual sets of 1 gene each.

Our work suggests the promise of projected gradient methods for other $\ell_1$-regularized problems by optimization in the dual space, as the duals of these problems often have simple $\ell_\infty$ constraints to which it is trivial to project. However, this approach currently relies on the ability to efficiently recover the primal variables from the dual variables, making its general application an open problem. Another interesting direction for future work is the construction of methods for intelligently setting the penalty parameters $\lambda$. This could certainly lead to more accurate structure recoveries, and recent work by Do et al. (2007) demonstrates the promise of hyperparameter learning in log-linear models; this might be extended to more general problems such as structure learning. Our work demonstrates the benefit of block-structured regularization. However, the blocks must currently be selected by hand and cannot overlap. Automatically learning the block structure would be a very useful extension to our work, both for improved performance and for the explanatory power it gives beyond individual edge structures. Finally, $\ell_1$-regularized learning has recently been demonstrated successfully for discrete MRFs (Lee et al., 2006; Wainwright et al., 2007). It would be interesting to see whether efficient projection-based methods such as ours can be applied to richer settings that involve discrete variables or non-linear continuous interactions.

**Acknowledgments**

We thank Stephen Boyd for his guidance and useful comments, Gal Elidan and Geremy Heitz for help with the mammals data set, Gal Chechik and Su-In Lee for help with the gene expression data set, and the anonymous reviewers for helpful feedback. This work was supported by the Office of Naval Research under MURI N000140710747, DARPA under the Transfer Learning program, and NSF grant BDI-0345474.